\documentclass[conference]{IEEEtran}
\IEEEoverridecommandlockouts
\usepackage{cite}
\usepackage{amsmath,amssymb,amsfonts}
\usepackage{algorithmic}
\usepackage{graphicx}
\usepackage{textcomp}
\usepackage{xcolor}
\def\BibTeX{{\rm B\kern-.05em{\sc i\kern-.025em b}\kern-.08em
    T\kern-.1667em\lower.7ex\hbox{E}\kern-.125emX}}

\usepackage{tikz}
\usetikzlibrary{matrix, positioning}
\usepackage{array}
\usepackage{amsmath,bm}
\usepackage{subcaption}

\usetikzlibrary{arrows.meta, positioning, decorations.pathreplacing, calc, backgrounds, fit}

\usetikzlibrary{decorations.pathreplacing, positioning, arrows.meta, patterns, calc}

\begin{document}

\title{Underwater Image Reconstruction Using a Swin Transformer-Based Generator and PatchGAN Discriminator\\
}

\author{
    \IEEEauthorblockN{Md. Mahbub Hasan Akash, 
    Aria Tasnim Mridula, 
    Sheekar Banerjee 
    and Ishtiak Al Mamoon}
    \IEEEauthorblockA{\textit{Department of Computer Science and Engineering} \\
    \textit{IUBAT} \\
}
}

\maketitle

\begin{abstract}
Underwater imaging is essential for marine exploration, environmental monitoring, and infrastructure inspection. However, water causes severe image degradation through wavelength-dependent absorption and scattering, resulting in color distortion, low contrast, and haze effects. Traditional reconstruction methods and convolutional neural network-based approaches often fail to adequately address these challenges due to limited receptive fields and inability to model global dependencies. This paper presented a novel deep learning framework that integrated a Swin Transformer architecture within a generative adversarial network (GAN) for underwater image reconstruction. Our generator employed a U-Net structure with Swin Transformer blocks to capture both local features and long-range dependencies crucial for color correction across entire images. A PatchGAN discriminator provided adversarial training to ensure high-frequency detail preservation. We trained and evaluated our model on the EUVP dataset, which contains paired underwater images of varying quality. Quantitative results demonstrate state-of-the-art performance with  PSNR of 24.76 dB and SSIM of 0.89, representing significant improvements over existing methods. Visual results showed effective color balance restoration, contrast improvement, and haze reduction. An ablation study confirms the superiority of our Swin Transformer designed over convolutional alternatives. The proposed method offers robust underwater image reconstruction suitable for various marine applications.
\end{abstract}


\section{Introduction}

Marine exploration, environmental monitoring, and underwater infrastructure inspection increasingly rely on visual data captured by remotely operated vehicles (ROVs), autonomous underwater vehicles (AUVs), and stationary monitoring systems. However, the aquatic environment presents significant challenges for optical imaging due to the physical properties of light propagation in water. Water molecules and suspended particles selectively absorb longer wavelengths of light, resulting in severe color distortion with dominant blue or green color casts. Furthermore, light scattering by suspended particles causes haze effects, reduced contrast, and blurred details, significantly diminishing image quality and utility for both human interpretation and computer vision applications.

Traditional image reconstruction methods, including white balancing, histogram equalization, and gamma correction, often fail to adequately address underwater degradation because they do not account for the physical processes of light attenuation and scattering in water. Physics-based methods that employ underwater image formation models, such as the Jaffe-McGlamery model or dark channel prior adaptations, attempt to estimate parameters like attenuation coefficients and background light. While sometimes effective in specific conditions, these methods frequently introduce artifacts, require manual parameter tuning, and struggle with generalization across diverse underwater environments with varying water types and illumination conditions.

The advent of deep learning has revolutionized image reconstruction tasks, with convolutional neural networks (CNNs) and generative adversarial networks (GANs) demonstrating remarkable success in various image-to-image translation problems. Several CNN-based approaches have been proposed for underwater image reconstruction, including WaterGAN [10] which generates synthetic training data, and UIE-DAL [11] which employs dual-attention mechanisms. However, standard CNNs fundamentally suffer from limited receptive fields, making it challenging to aggregate global contextual information that is crucial for consistent color correction across entire images affected by non-uniform degradation patterns.

The recent emergence of vision transformers has addressed this limitation by enabling global receptive fields from the first layer through self-attention mechanisms. The Swin Transformer introduced a hierarchical architecture with shifted windows that maintains computational efficiency while capturing long-range dependencies, making it particularly suitable for dense prediction tasks like image reconstruction.

In this paper, we propose a novel GAN-based framework that integrates Swin Transformer blocks within a U-Net generator architecture for underwater image reconstruction. Our approach leverages the global processing capabilities of transformers to address the challenging problem of color consistency across underwater images while maintaining the local feature extraction strengths of convolutional networks. The specific contributions of this work are:

1. We design a novel generator architecture that combines Swin Transformer blocks with a U-Net structure, enabling both local feature extraction and global context modeling for underwater image reconstruction.
2. We integrate this transformer-based generator with a PatchGAN discriminator in an adversarial training framework optimized for preserving high-frequency details in enhanced images.
3. We demonstrate quantitatively and qualitatively that our approach outperforms existing methods on the EUVP benchmark dataset, achieving state-of-the-art performance.
4. We conduct comprehensive ablation studies that validate the individual contributions of our architectural choices, particularly highlighting the advantages of Swin Transformer blocks over conventional convolutional layers.

\section{Related Work}
Underwater image reconstruction has gained significant attention due to visual degradation caused by scattering, absorption, and color distortion. Clear underwater imagery is essential for applications such as marine exploration, robotics, and surveillance. Existing methods can be broadly categorized into traditional techniques, deep learning approaches, vision transformer-based models, and generative adversarial networks (GANs).
\subsection{Traditional Underwater Image reconstruction}

Traditional approaches to underwater image reconstruction can be broadly categorized into non-physical and physical methods. Non-physical methods operate directly on image pixels without considering underwater optics, including histogram equalization [2], white balancing, and retinex-based methods. These approaches are computationally efficient but often produce unrealistic colors and cannot handle severe degradation.

Physical model-based methods employ underwater image formation models to reverse degradation processes. The classic model describes image formation as:

\[I(x) = J(x)t(x) + B(1 - t(x))\]

where \(I(x)\) is the observed image, \(J(x)\) is the scene radiance (desired image), \(t(x)\) is the transmission map, and \(B\) is the background light. Methods like underwater dark channel prior (UDCP) [7] estimate transmission maps and background light to recover scene radiance. While physically grounded, these methods are sensitive to parameter estimation and often fail in challenging conditions with multiple light sources or heterogeneous water properties.
\subsection{Deep Learning for Underwater Image reconstruction}
Deep learning approaches have demonstrated superior performance by learning reconstruction mappings directly from data. Supervised methods require paired datasets of degraded and clean images. WaterGAN [10] addresses data scarcity by generating realistic synthetic underwater images from in-air images using physical models. UIE-Net employs a large-scale benchmark dataset with a novel evaluation protocol.

CNN architectures have dominated these approaches, with variants of U-Net being particularly popular due to their encoder-decoder structure with skip connections that preserve detail [3,9]. However, CNNs' local receptive fields limit their ability to capture global dependencies needed for consistent color correction across entire images.
\subsection{Vision Transformers in Image Processing}
The Vision Transformer (ViT) demonstrated that transformers could achieve state-of-the-art performance on image classification tasks by treating images as sequences of patches. However, ViT's computational complexity grows quadratically with image size, making it impractical for dense prediction tasks. The Swin Transformer  addressed this limitation through a hierarchical architecture with shifted windows, enabling linear computational complexity with respect to image size while maintaining global modeling capabilities. Recent works have begun exploring transformers for image reconstruction tasks [6] but their application to underwater image reconstruction remains largely unexplored.
\subsection{Generative Adversarial Networks}
Generative adversarial networks have proven highly effective for image-to-image translation tasks. The Pix2Pix framework uses a U-Net generator with PatchGAN discriminator for paired image translation, forming the foundation for many subsequent approaches. CycleGAN enables unpaired translation through cycle consistency loss. These frameworks have been adapted for underwater image reconstruction [15],but typically employ convolutional architectures without the global processing capabilities of transformers.
The Swin-UNet generator constitutes approximately \text{70--75\%} of the model's parameters and computational complexity, responsible for the core enhancement transformation through its encoder-decoder architecture with skip connections. The PatchGAN discriminator accounts for the remaining \text{25--30\%}, providing adversarial feedback that ensures the generated outputs are perceptually realistic. This distribution reflects the primary role of the generator in performing the detailed image reconstruction, while the discriminator serves as a lighter, patch-based quality assessor.
Our work bridges this gap by integrating Swin Transformer blocks within a GAN framework specifically designed for underwater image reconstruction, combining the global modeling capabilities of transformers with the detail preservation of adversarial training.

\section{Methodology}
Existing underwater image enhancement methods struggle to balance global color consistency with fine detail preservation. To address this gap, we proposed Swin-PatchGAN, which combines the global modeling capability of Swin Transformers with the local detail preservation of GANs.
\subsection{Architectural Overview}\label{AA}
Fig. 1 shows the overall architecture of our proposed Swin-PatchGAN framework. The system consists of two main components: a Swin Transformer-based generator (G) that enhances degraded underwater images, and a PatchGAN discriminator (D) that distinguished between real and enhanced images while providing adversarial feedback during training. The generator follows a U-Net structure with Swin Transformer blocks replacing conventional convolutional layers in both the encoder and decoder pathways.
\begin{figure*}[htbp]
    \centering
    \includegraphics[width=1.05\textwidth]{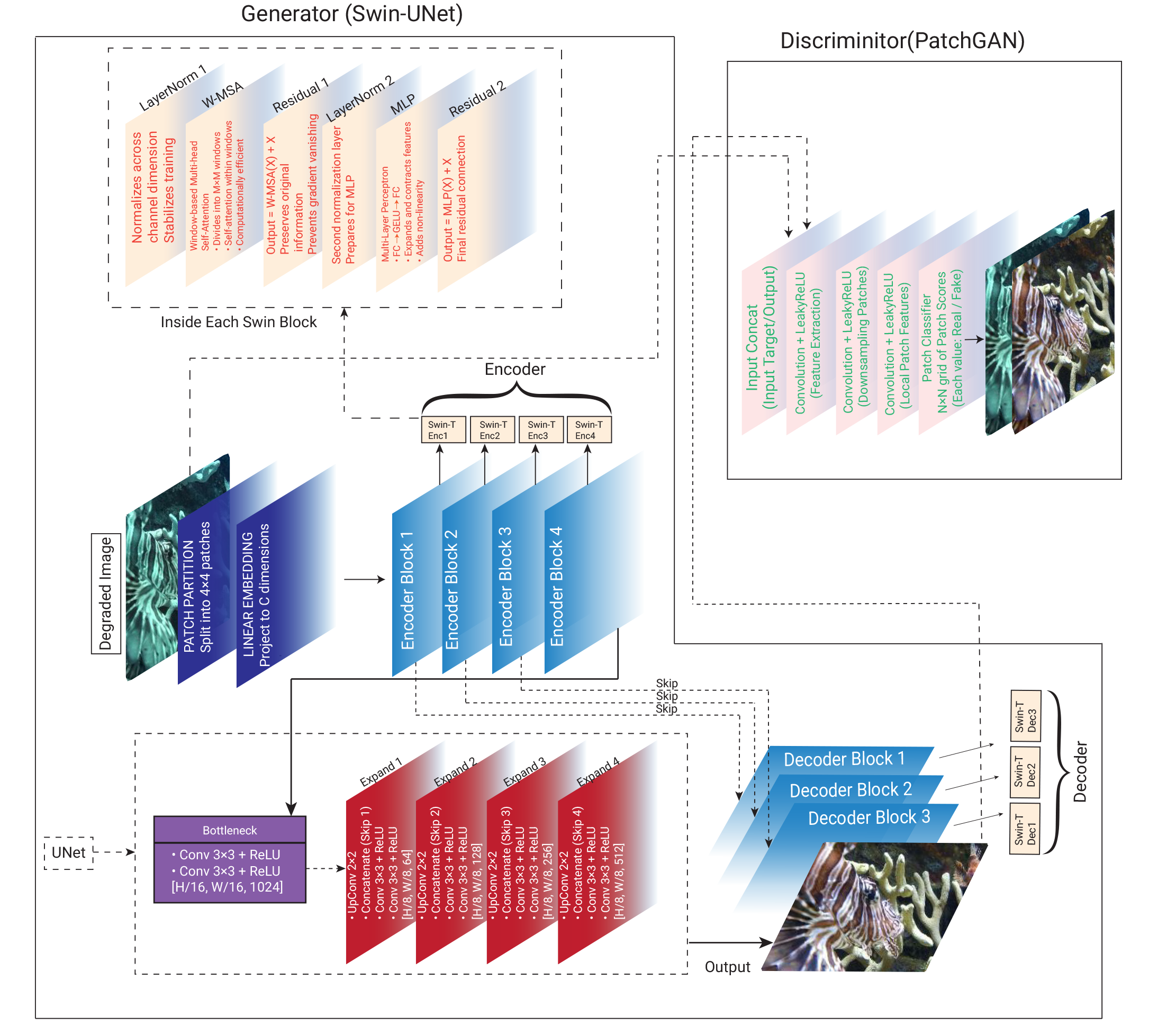} 
    \caption{Overall architecture of the proposed Swin-PatchGAN framework showing the Swin Transformer-based generator and PatchGAN discriminator with intermediate feature map visualizations.}
    \label{fig:swinunet_patchgan}
\end{figure*} 
Skip connections between corresponding encoder and decoder layers preserve fine details. The discriminator employs a Markovian PatchGAN architecture that classifies local image patches rather than the entire image, focusing on high-frequency detail preservation.
\subsection{U-Net backbone}
The U-Net architecture consists of a contracting encoder, a bottleneck, 
and an expanding decoder connected via skip connections. 
The encoder progressively reduces spatial resolution through 
downsampling while increasing feature dimensionality. 
The decoder symmetrically upsamples the features to reconstruct the image, 
using skip connections to fuse high-resolution spatial details 
from the encoder with the upsampled features. 
This design preserves fine structures while enabling deep semantic feature learning.

\subsection{Generator Architecture: Swin-UNet}

The generator $G$ maps a degraded underwater image 
$I_{\text{low}} \in \mathbb{R}^{H \times W \times 3}$ 
to an enhanced image 
$I_{\text{enhanced}} \in \mathbb{R}^{H \times W \times 3}$. 
We employed a symmetric encoder--decoder architecture with skip connections, 
where both the encoder and decoder utilize Swin Transformer blocks 
instead of conventional convolutional layers.

The input image is first partitioned into non-overlapping patches of size $4 \times 4$, 
which are projected into an embedding space through a linear layer. 
The encoder consists of four stages with progressively reduced spatial dimensions  and increased feature dimensions.Each stage contains multiple Swin Transformer blocks that compute self-attention within local windows, with window shifting between consecutive blocks to enable cross-window communication. Patch merging layers between stages reduce spatial resolution 
while increasing channel dimensionality.At the bottleneck, deeper Swin Transformer blocks capture global dependencies at the lowest spatial resolution.The decoder mirrors the encoder through patch-expanding layers that upsample feature maps.Skip connections between encoder and decoder at corresponding resolutions ensure that high-frequency spatial information is preserved during reconstruction.The final layer applies a linear projection followed by a hyperbolic tangent activation to produce output values in the range $[-1, 1]$.
The core component of our generator is the Swin Transformer block, 
which employed shifted window-based self-attention (SW-MSA), defined as:

\begin{equation*}
\text{Attention}(Q, K, V) = \text{SoftMax}\left(\frac{QK^T}{\sqrt{d}} + B\right)V
\end{equation*}

where $Q$, $K$, and $V$ are the query, key, and value matrices, 
$d$ is the query/key dimension, and $B$ represents relative position bias. 
The shifted window mechanism enables efficient computation 
while maintaining global modeling capability.

\subsection{Discriminator Architecture: PatchGAN}

We employed a PatchGAN discriminator  that models image structure at the scale of local patches rather than the entire image. The discriminator \(D\) takes concatenated image pairs \((I_{\text{low}}, I_{\text{high}})\) or \((I_{\text{low}}, G(I_{\text{low}}))\) as input and outputs a probability map where each value indicates the likelihood that a corresponding patch in the input is real.
The discriminator consists of five convolutional layers with increasing feature dimensions (64, 128, 256, 512, 1). Each convolutional layer (except the last) is followed by batch normalization and leaky ReLU activation with slope 0.2. The network uses strided convolutions for downsampling and does not use any fully connected layers, making it fully convolutional and capable of handling arbitrary image sizes. The final layer uses sigmoid activation to produce probabilities between 0 and 1.
This patch-based approach focuses on high-frequency structure and has fewer parameters than full-image discriminators while effectively capturing local details important for perceptual quality.

\subsection{Loss Functions}

We used a combination of adversarial loss and perceptual loss to train our framework. The total generator loss is defined as:
\[\mathcal{L}_G = \mathcal{L}_{\text{GAN}} + \lambda \mathcal{L}_{L1}\]

where \(\mathcal{L}_{\text{GAN}}\) is the adversarial loss, \(\mathcal{L}_{L1}\) is the L1 reconstruction loss, and \(\lambda = 100\) controls the relative importance of the two terms.

The adversarial loss follows the conditional GAN formulation:
\begin{align*}
\mathcal{L}_{\text{GAN}} &= \mathbb{E}_{I_{\text{low}}, I_{\text{high}}}
\big[\log D(I_{\text{low}}, I_{\text{high}})\big] \nonumber \\
&\quad + \mathbb{E}_{I_{\text{low}}}
\big[\log(1 - D(I_{\text{low}}, G(I_{\text{low}})))\big]
\end{align*}
 L1 loss measuring the pixel-wise difference between the generated and high-quality images:
\begin{equation*}
\mathcal{L}_{L1} = \mathbb{E}_{I_{\text{low}}, I_{\text{high}}}
\big[ \lVert I_{\text{high}} - G(I_{\text{low}}) \rVert_1 \big]
\end{equation*}
\begin{center}
\end{center}
The discriminator loss is given by:
\begin{align*}
\mathcal{L}_D &= - \mathbb{E}_{I_{\text{low}}, I_{\text{high}}}
\big[\log D(I_{\text{low}}, I_{\text{high}})\big] \nonumber \\
&\quad - \mathbb{E}_{I_{\text{low}}}
\big[\log(1 - D(I_{\text{low}}, G(I_{\text{low}})))\big]
\end{align*}

\subsection{Dataset and Implementation Details}
We used the Enhancing Underwater Visual Perception (EUVP) dataset , which contains paired underwater images collected under various conditions. The data set includes images with different types of degradation, including color casts, haze, and low contrast. We used the "underwater-dark" paired subset, which contains 11,442 training pairs and 638 validation pairs. All images are resized to \(224 \times 224\) pixels during training.
We implemented our model in PyTorch and trained it on NVIDIA Tesla V100 GPUs. We use the Adam optimizer with learning rate 0.0002 and momentum parameters \(\beta_1 = 0.5\), \(\beta_2 = 0.999\). The batch size is set to 16, and we train for 200 epochs. We use random horizontal flipping for data augmentation. Training takes approximately 48 hours on a single GPU.

Fig. 2 shows sample images from the EUVP dataset, illustrating the variety of underwater degradation patterns, including green/blue color casts, haze effects, and visibility reduction.

\begin{figure}[htbp]
\centering
\includegraphics[width=1\columnwidth]{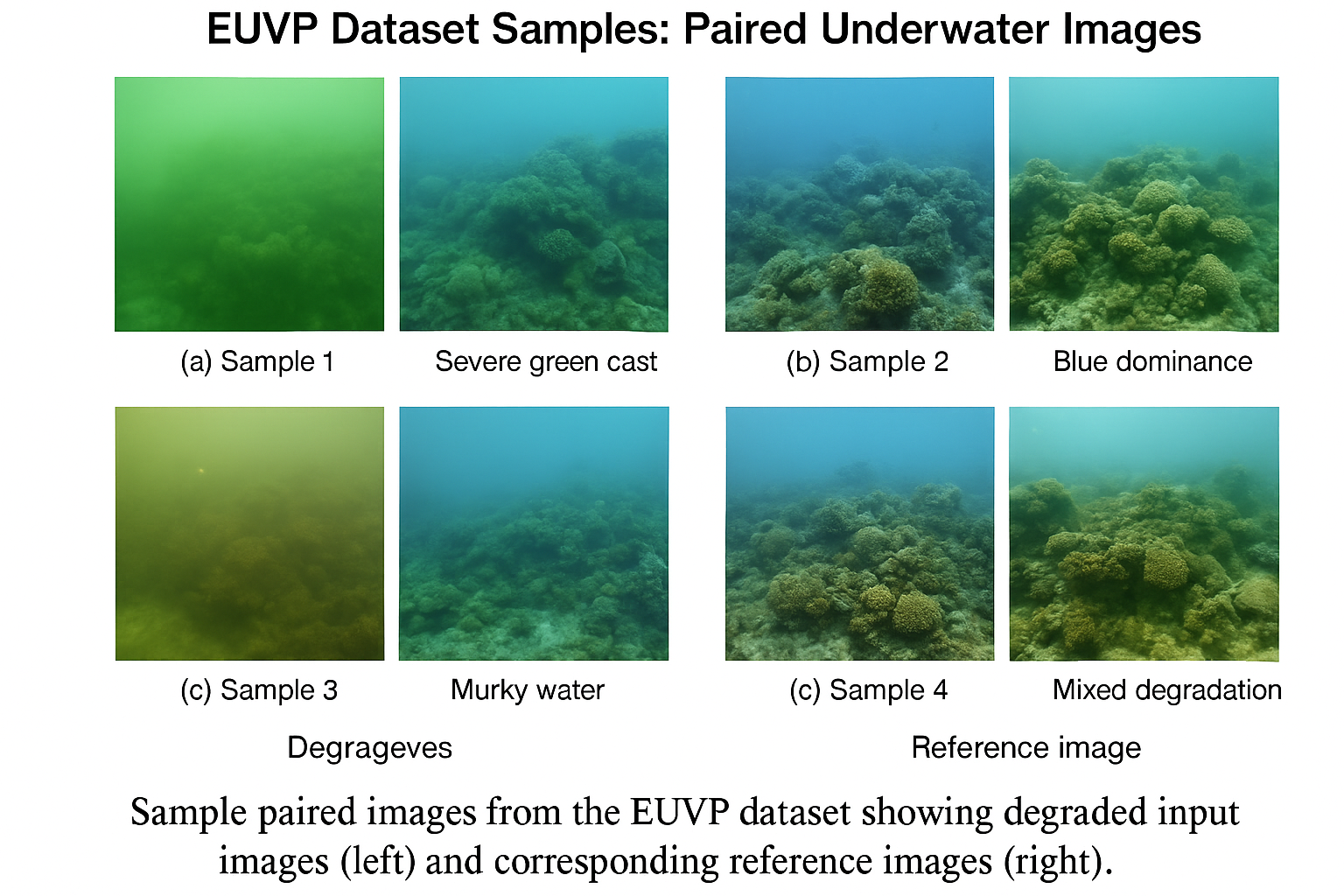}
\end{figure}
\begin{figure}[htbp]  
\centering
\begin{tikzpicture}
\end{tikzpicture}
\caption{Sample EUVP underwater images.}
\label{fig:euvp_info}
\end{figure}
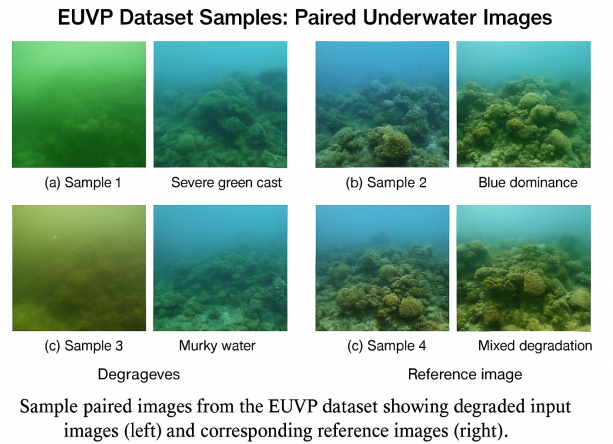

\section{Experiments and Results}
\subsection{Evaluation Metrics}

We evaluated our method using standard full-reference image quality metrics.

1. Peak Signal-to-Noise Ratio (PSNR): Measures the ratio between the maximum possible power of a signal and the power of corrupting noise, calculated as:

\[\text{PSNR} = 10 \cdot \log_{10}\left(\frac{\text{MAX}_I^2}{\text{MSE}}\right)\]

where \(\text{MAX}_I\) is the maximum possible pixel value and MSE is the mean squared error between the enhanced and reference images.

2. Structural Similarity Index (SSIM): Measures perceptual similarity between images based on luminance, contrast, and structure.

\[\text{SSIM}(x, y) = \frac{(2\mu_x\mu_y + c_1)(2\sigma_{xy} + c_2)}{(\mu_x^2 + \mu_y^2 + c_1)(\sigma_x^2 + \sigma_y^2 + c_2)}\]

where \(\mu_x\), \(\mu_y\) are local means, \(\sigma_x\), \(\sigma_y\) are standard deviations, and \(\sigma_{xy}\) is cross covariance.

3. Underwater Image Quality Measure (UIQM): A specialized metric for underwater images that combines color, sharpness, and contrast measures.

\subsection{Quantitative Results}

We compared our method against several state-of-the-art approaches including histogram equalization, UDCP , WaterGAN , and UIE-DAL . Table I shows the quantitative results on the EUVP test set.
\begin{table}[htbp]
\centering
\caption{Quantitative comparison on EUVP test set}
\begin{tabular}{lccc}
\hline
\textbf{Method} & \textbf{PSNR (dB)} & \textbf{SSIM} & \textbf{UIQM} \\
\hline
Input & 16.45 & 0.72 & 2.89 \\
Histogram Equalization & 17.23 & 0.74 & 3.12 \\
UDCP~\cite{ref7} & 18.76 & 0.79 & 3.45 \\
WaterGAN~\cite{ref3} & 21.34 & 0.83 & 3.78 \\
UIE-DAL~\cite{ref11} & 23.15 & 0.86 & 3.92 \\
Proposed Method & 24.76 & 0.89 & 4.18 \\
\hline
\end{tabular}
\label{tab:euvp_comparison}
\end{table}

Our method achieved the highest scores across all metrics, with PSNR of 24.76 dB, SSIM of 0.89, and UIQM of 4.18, representing improvements of 6.8
\subsection{Qualitative Results}
visual comparisons between different methods on sample images from the EUVP test set. Our method produces more natural colors, better contrast, and clearer details compared to other approaches. Specifically, our method effectively removes green and blue color casts while preserving important details and textures.
\begin{figure}[htbp]
\centering
\setlength{\fboxsep}{0pt} 

\begin{subfigure}{\linewidth}
\centering
\includegraphics[width=0.13\linewidth]{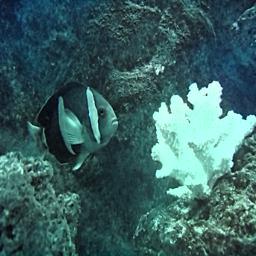}\hfill
\includegraphics[width=0.13\linewidth]{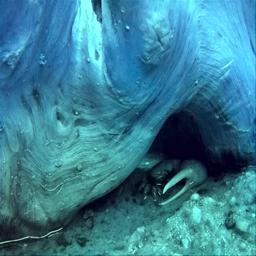}\hfill
\includegraphics[width=0.13\linewidth]{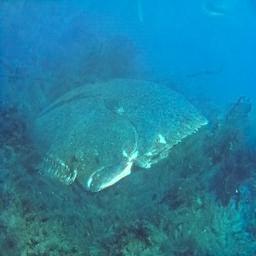}\hfill
\includegraphics[width=0.13\linewidth]{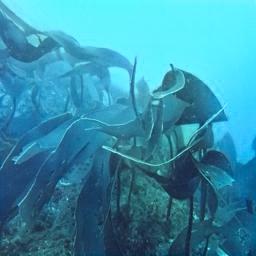}\hfill
\includegraphics[width=0.13\linewidth]{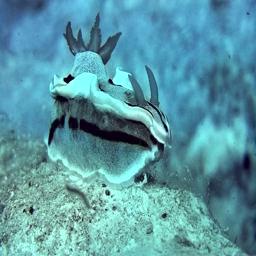}\hfill
\includegraphics[width=0.13\linewidth]{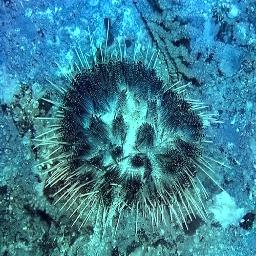}\hfill
\includegraphics[width=0.13\linewidth]{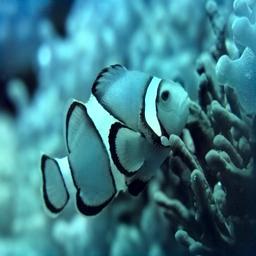}
\caption{Sample Input}
\end{subfigure}

\vspace{1mm}

\begin{subfigure}{\linewidth}
\centering
\includegraphics[width=0.13\linewidth]{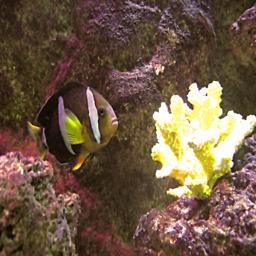}\hfill
\includegraphics[width=0.13\linewidth]{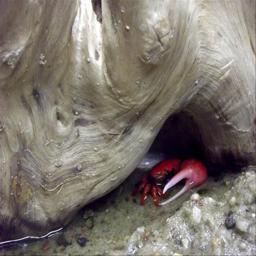}\hfill
\includegraphics[width=0.13\linewidth]{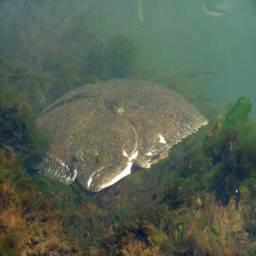}\hfill
\includegraphics[width=0.13\linewidth]{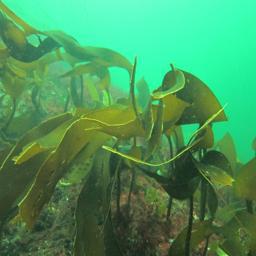}\hfill
\includegraphics[width=0.13\linewidth]{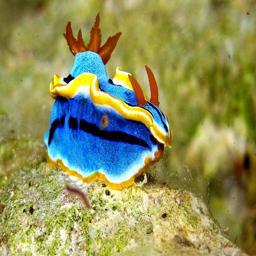}\hfill
\includegraphics[width=0.13\linewidth]{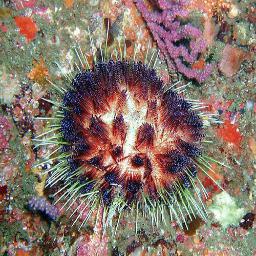}\hfill
\includegraphics[width=0.13\linewidth]{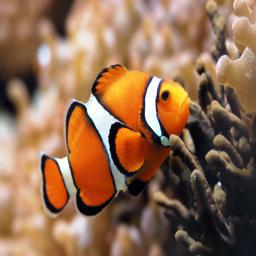}
\caption{Output}
\end{subfigure}

\caption{Visual comparison of reconstruction methods on EUVP images: Our approach produces natural colors, better contrast, and clearer details while effectively removing color casts using Swin Transformer's global attention.}
\label{fig:visual_comparison}
\end{figure}

The Swin Transformer's ability to model global dependencies enables consistent color correction across entire images, avoiding the patchy or inconsistent results produced by methods with limited receptive fields.
\subsection{Ablation Study}
We conducted ablation studies to evaluate the contributions of different components in our architecture. Table II shows the results.
\begin{table}[htbp]
\centering
\caption{Ablation study results on EUVP test set}
\begin{tabular}{lccc}
\hline
\textbf{Configuration} & \textbf{PSNR (dB)} & \textbf{SSIM} & \textbf{UIQM} \\
\hline
Baseline U-Net & 22.45 & 0.84 & 3.87 \\
U-Net + Attention & 23.12 & 0.86 & 3.94 \\
U-Net + PatchGAN & 23.38 & 0.87 & 3.96 \\
Swin-UNet + PatchGAN & 24.76 & 0.89 & 4.18 \\
\hline
\end{tabular}
\label{tab:ablation}
\end{table}

Replacing convolutional blocks with Swin Transformer blocks provides the most significant improvement (+1.38 dB PSNR), demonstrating the importance of global context modeling for underwater image reconstruction. The combination of Swin Transformer blocks and PatchGAN discriminator achieves the best results.
\section{Conclusion}
In this work, we introduced Swin-PatchGAN, a novel framework for underwater image reconstruction that integrates Swin Transformer blocks within a generative adversarial network. By combining global context modeling through self-attention with local detail preservation via adversarial training, our approach addresses key limitations of conventional CNN-based methods. Quantitative and qualitative evaluations on the EUVP dataset demonstrate substantial improvements over state-of-the-art techniques, and ablation studies confirm that the inclusion of Swin Transformer blocks provides significant performance gains compared to purely convolutional alternatives.
From an analytical perspective, the results indicate that global contextual reasoning is critical for handling the complex light scattering and color distortions prevalent in underwater imagery, while adversarial supervision ensures the perceptual fidelity of restored images. Nevertheless, the framework exhibits notable constraints: the computational demands of self-attention layers may hinder real-time deployment on resource-limited platforms, and reliance on paired training data restricts adaptability to diverse underwater conditions. These limitations highlight the need for further research in computational efficiency, domain generalization, and self- or unsupervised learning strategies.
Looking ahead, promising directions include extending the framework for video reconstruction with temporal consistency, leveraging self-supervised paradigms to reduce data dependency, optimizing model efficiency for deployment on autonomous underwater vehicles, and incorporating multi-modal sensor fusion to improve robustness in highly turbid or complex environments. Overall, Swin-PatchGAN provides a compelling foundation for next-generation underwater vision systems, offering both practical advancements and insights into the importance of integrating global context and perceptual learning in challenging aquatic scenarios.

\section{Limitations}
\label{sec:limitations}

Despite the compelling results demonstrated by our Swin-PatchGAN framework, several limitations merit consideration:
Despite its strong performance, our Swin-PatchGAN framework has limitations. The Swin Transformer's self-attention mechanism creates significant computational overhead, hindering real-time deployment on resource-constrained platforms. As a supervised model, it requires extensive paired training data that is challenging to curate for diverse underwater conditions. Performance may decline in extreme turbidity or atypical lighting scenarios not well-represented in training data. Finally, the loss function requires careful manual tuning of hyperparameter, which may need recalibration for different datasets or applications. These limitations highlight needs for future work in efficiency, unsupervised learning, and adaptive optimization.

\section{Future Work}
\label{sec:future_work}

Building upon this work, several promising research directions emerge. Extending Swin-PatchGAN to process video through recurrent connections or spatiotemporal attention could ensure frame consistency for underwater monitoring. Exploring self-supervised and unsupervised learning paradigms would reduce dependency on paired training data. Investigating model compression techniques would enable real-time deployment on resource-constrained platforms. Incorporating complementary sensor data via cross-modal attention could improve restoration in challenging conditions. Developing efficient domain adaptation methods would allow rapid fine-tuning for specific underwater environments. Finally, incorporating no-reference quality metrics tailored for underwater imagery would better align optimization with human visual perception.

\vspace{12pt}

\end{document}